\documentclass[journal]{IEEEtran}

\usepackage{amsfonts}
\usepackage[cmex10]{amsmath}
\usepackage{amsbsy}
\usepackage{amssymb}
\usepackage{amsthm}
\usepackage{graphicx}
\usepackage{caption}
\usepackage{subcaption}
\usepackage{epsf}
\usepackage{algorithm}
\usepackage{algorithmic}
\usepackage{multirow}
\usepackage{mathrsfs}
\usepackage{amsmath}
\usepackage{bm}
\usepackage{comment}

\newcounter{RomanNumber}

\DeclareGraphicsExtensions{.eps,.pdf,.jpg,.png}
\DeclareMathAlphabet{\mathsfsl}{OT1}{cmss}{m}{sl}
\usepackage{color}

\ifCLASSINFOpdf
\else
\fi
\begin{document}
\title{Anomaly Detection with Generative Adversarial Networks for Multivariate Time Series}
\author{Dan~Li,~
        Dacheng~Chen,~
        Jonathan~Goh,~
        and~See-Kiong~Ng,~

\thanks{This work was presented in the 7th International Workshop on Big Data, Streams and Heterogeneous Source Mining: Algorithms, Systems, Programming Models and Applications on the ACM Knowledge Discovery and Data Mining conference, August 2018, London, United Kingdom}
\thanks{Dan Li , Dacheng Chen and See Kiong Ng are with the Institute of Data Science, National University of Singapore, 3 Research Link, Singapore 117602 
}
\thanks{Jonathan Goh is with the ST Electronics (Info Security) Pte Ltd. 
}
\thanks{The code is available at https://github.com/LiDan456/GAN-AD}
}

\maketitle

\begin{abstract}
Today's Cyber-Physical Systems (CPSs) are large, complex, and affixed with networked sensors and actuators that are targets for cyber-attacks. Conventional detection techniques are unable to deal with the increasingly dynamic and complex nature of the CPSs.  On the other hand, the networked sensors and actuators generate large amounts of data streams that can be continuously monitored for intrusion events.   Unsupervised machine learning techniques can be used to model the system behaviour and classify deviant behaviours as possible attacks. In this work, we proposed a novel Generative Adversarial Networks-based Anomaly Detection (GAN-AD) method for such complex networked CPSs. We used LSTM-RNN in our GAN to capture the distribution of the multivariate time series of the sensors and actuators under normal working conditions of a CPS.  Instead of treating each sensor's and actuator's time series independently, we model the time series of  multiple sensors and actuators in the CPS concurrently to take into account of potential latent interactions between them.  To exploit both the generator and the discriminator of our GAN, we deployed the GAN-trained discriminator together with the residuals between generator-reconstructed data and the actual samples to detect possible anomalies in the complex CPS. We used our GAN-AD to distinguish abnormal attacked situations from normal working conditions for a complex six-stage Secure Water Treatment (SWaT) system. Experimental results showed that the proposed strategy is effective in identifying anomalies caused by various attacks with high detection rate and low false positive rate as compared to existing methods.
\end{abstract}


%
\IEEEpeerreviewmaketitle

\section{Introduction}
\label{sec:Intro}

Cyber-Physical Systems (CPSs) are interconnected physical systems typically engineered for mission-critical tasks.  Some example CPSs are water treatment and distribution plants, natural gas distribution systems, oil refineries, power plants, power grids, and autonomous vehicles. The emergence of the Internet of Things (IoT) will further drive the proliferation of CPSs for a large variety of tasks, resulting in many systems and devices communicating and operating autonomously over networks. As such, cyber-attacks are one of the most concerned potential threats to CPSs.  

Traditionally, Statistical Process Control (SPC) methods such as CUSUM, EWMA and Shewhart charts are popular solutions for anomaly detection \cite{Sun:2014}. These conventional detection techniques are unable to deal with the increasingly dynamic and complex nature of the CPSs  with the advent of IoT. As such, researchers have moved beyond specification or signature-based techniques and begun to exploit both supervised and unsupervised machine learning techniques to develop more intelligent and adaptive methods from big data to identify anomalies or intrusions \cite{Donghwoon2017}. 

However, even with the use of machine learning techniques, detecting anomalies in time series is still challenging.  First, most of the supervised techniques require enough liable normal data and labelled anomaly classes to learn from but this is hardly the case in practice as anomalies are typically rare.  Secondly, most of existing unsupervised methods are built through linear projection and transformation but there is often non-linearity in the hidden inherent correlations of the multivariate time series of complex CPSs.  Most of the current techniques also employ simple comparison between present state and predicted normal ranges, which can be inadequate for anomaly detection since the control bounds are not flexible enough and cannot effectively identify indirect attacks\footnote{For one specific variable (i.e. a sensor or actuator in the CPS), a "direct attack" is defined as an attack that is directly inserted to it and affects its performance, while an "indirect attack" is an intrusion targeted for another variable but also affects the performance of the variable.}. 

\begin{figure*}[!t]
\centering
\includegraphics[width=1.0\textwidth]{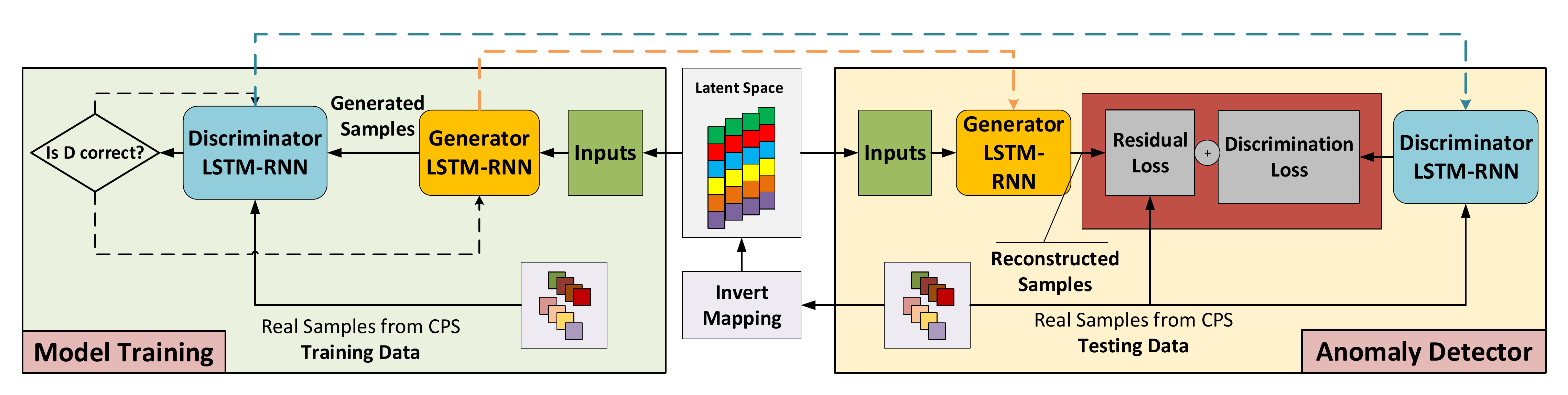}
\vspace{-0.5cm}
\caption{GAN-AD: Unsupervised GAN-based anomaly detection for CPSs. On the left is a GAN framework in which the generator and discriminator are obtained with iterative adversarial training. On the right is the exploitation of both the GAN's generator and discriminator for anomaly detection---the  generator is used for computing the residual loss between reconstructed samples and real ones, while the discriminator is used to compute the discrimination loss.}
\label{fig:GAN-CPS}
\end{figure*}

To address these challenges, we propose a novel unsupervised GAN-based Anomaly Detection (GAN-AD) method for a complex multi-process CPS with multiple networked sensors and actuators by modelling the non-linear correlations among multiple time series and detecting anomalies based on the trained GAN model.  Fig. \ref{fig:GAN-CPS} depicts the overall framework of our proposed GAN-AD.   First, to deal with time-series data, the generator and discriminator are built as two Long-Short-Term Recurrent Neural Networks (LSTM-RNN), as shown in the left part of Fig. \ref{fig:GAN-CPS}. As in a typical GAN,  the generator (G) generates fake samples from a specific latent space and passes that to the discriminator (D) which tries to distinguish fake from real. Based on the outputs of D, the system will update parameters of D and G, so that the discriminator will be trained to be as sensitive as possible to assign correct labels to both real and fake samples, while the generator will be trained as smart as possible to fool the discriminator (assigning real labels to fake samples). After sufficient rounds of iterations, the generator will have captured the hidden distributions of the training sequences and that could generate realistic samples.  In other words, G can be viewed as an implicit model of the CPS.  At the same time, the resulting discriminator D is also able to distinguish fake from real with high sensitivity.  In other words, D is an intuitive tool for anomaly detection. In this work, we propose to exploit both G and D for the anomaly detection task by  (i) exploiting the residuals between real-time testing samples and reconstructed samples based on the mapping from real-time space to the GAN latent space; and (ii) discrimination with the machine-learned discriminator by classifying the real-time series. We depict this aspect of our proposed GAN-AD in the right part of Fig. \ref{fig:GAN-CPS}.  As shown, the testing samples are mapped back into the latent space, and  the corresponding residual loss is calculated based on the difference between the reconstructed testing samples (by the trained generator) and the actual testing samples. At the same time, testing samples are also fed to the trained discriminator to compute the discrimination loss. The two losses are then combined to detect potential anomalies for sequential CPS data (more details are described in Section \ref{subsec:AD}). 

The remaining part of this paper is organized as follows. Section \ref{sec:RW} introduces the related works. Section \ref{sec:Model} presents our proposed GAN-AD and derives an anomaly score function. Section \ref{sec:CPSandAtt} introduces the multi-stage Secure Water Treatment system, which is followed by Section \ref{sec:ExpandRes} in which we evaluate our proposed GAN-AD on real-time multivariate series. Finally, Section \ref{sec:ConcandFW} summarizes the whole paper and proposes possible future work.

\section{Related Works}
\label{sec:RW}

The basic task of anomaly detection is to identify whether the testing data conform to the normal data distribution; the non-conforming points are  called anomalies, outliers, intrusions, failures or contaminants in various application domains \cite{Chandola2012, Donghwoon2017}. Anomaly detection is an old but challenging problem---it has been studied in the statistics community as early as the 19th century \cite{Chandola2012}. 

Based on how the historical training data is used, we can broadly divide anomaly detection methods into three categories: i) Statistical Process Control (SPC) techniques, ii) supervised machine learning methods, and iii) unsupervised machine learning methods.  The SPC techniques were extensively used in the early years for monitoring and controlling quality of manufacturing processes through univariate or multivariate analysis \cite{YeNong2003}. The SPC approaches typically inspect changes in process mean (mean shifts) and process variance (variance changes), and try to model the relationship among multiple variables \cite{Ryan2011,Montgomery2009}. Shewhart control charts and CUSUM control charts are univariate SPC techniques that are usually appied to detect mean shifts \cite{Roberts1959, Montgomery1991}. EWMV control charts are sensitive to univarite variance changes as well as mean shifts \cite{Chao-WenLu1999}. Although widely used, the aforementioned SPC techniques cannot model the correlation among various sequences, while interrelations are common even in traditional manufacturing systems. As an improvement, Hotelling's $T_2$ \cite{Miller1998}, Multivariate Cumulative Sum (MCUSUM) \cite{Woodall1985} and Multivariate Exponentially Weighted Moving ASverage (MEWMA) \cite{Lowry1992} were proposed to monitor the performance of multiple variables in a manufacturing system. However, these multivariate methods require independent and identically distributed (iid) assumption which is often violated in reality \cite{YeNong2001}. Moreover, the computationally intensive multivariate SPC methods are not practical for modern CPSs with high  complexity and massive data streams.

Supervised machine learning techniques can also be used for anomaly detection. A typical approach is to build a predictive classification model for the normal and the anomalous classes. The classification model is trained from the labelled data.  New measurements can then be analysed by the classifier and be classified to corresponding categories (normal or anomalous) automatically \cite{Koturwar2015}. A wide range of supervised machine learning tecniques has  been used for anomaly detection.  They include Multivariate Regression models \cite{Mustafaraj:2010}, Bayes Classifier \cite{Xiao:2014}, Neural Networks (NN) \cite{Du:2014}, Fisher Discriminant Analysis (FDA) \cite{Dan1:2016}, Gaussion Mixture Model \cite{Jaikumar:2011}, Support Vector Data Description (SVDD) \cite{Zhao:2014}, Support Vector Machines (SVM) \cite{Mulumba:2015}, and tree-structured learning method \cite{Dan2:2016,Dan3:2016}. However, supervised classification methods are dependent on the availability of initial labelled training data.  Given that anomalies are typically rare, obtaining enough accurate labelled anomaly classes is usually challenging. 

The unsupervised learning methods---also known as descriptive or undirected classification---train models without lablled classes. Due to their simplicity and ability to handle large number of process data, unsupervised learning methods have been wildly used for anomaly detection for various industrial processes \cite{YinSD2014}. Popular unsupervised methods include Principal Component Analysis (PCA) \cite{Li:2014} and Partial Least Squares (PLS) \cite{HeXiao2013}. PCA is a multivariate data analysis method which preserves the significant variability information extracted from the process measurements and reduces the dimension for huge amount of correlated data \cite{Wold1987}. PLS is another multivariate data analysis method that has been extensively utilized for model building and anomaly detection \cite{Herman1985}. Their key performance indicators (Square Predicted Error (SPE) for PCA and $T^2$ index for PLS) for anomaly detection can be achieved with  correlation model traine off-line  and online process measurements. However, these unsupervised methods are only effective to highly correlated data, and require the data to follow multivariate Gaussian distribution \cite{XuewuD2013}. 

The recently proposed GAN framework enables researchers to build a generative model via adversarial training \cite{Goodfellow2014}. The simultaneous training of a generator and a discriminator in an adversarial fashion is highly suggestive for using the GAN framework for anomaly detection. The current successes of GANs are mainly in generating realistic-looking images.  In our CPS and IoT scenarios, we have to deal with oftentimes multiple streams of potentially interacting time series data.  However,there has been limited work in adopting the GAN framework for time-series data todate. To the best of our knowledge, there are two preliminary work using GAN to generate continuous valued sequences in the literature---one to produce polyphonic music with recurrent neural networks as generator and discriminator \cite{Mogren2016}, and the other uses conditional version of recurrent GAN to generate real-valued medical time series \cite{Esteban2017}. In both of these works, the multi-sequences were treated as i.i.d. and fed to a uniform GAN framework.  This will be inadequate for the IoT and CPSs setting given the potential interactions amongst the multiple time series-generating sensors and actuators involved in the same or different processes of the complex systems. 

Unlike traditional classification methods, the GAN-trained discriminator learns to detect false from real in an unsupervised fashion, making GAN an attractive unsupervised machine learning technique for anomaly detection \cite{XueY2017}. In addition, the GAN framework also produces a generator which is actually an inexplicit model of the target system with its ability to output normal samples from a certain latent space. Inspired by \cite{Raymond2016} and \cite{Salimans2016} that updates a mapping from the real-time space to a certain latent space to enhance the training of generator and discriminator, researchers have recently proposed to train a latent space understandable GAN and apply it for unsupervised learning of rich feature representations for arbitrary data distributions. \cite{Schlegl2017} and \cite{Zenati2018} showed the possibility of recognizing anomalies with reconstructed testing samples from latent space, and successfully applied the proposed GAN-based detection strategy to discover unexpected markers for images.  In this work, we will leverage on these previous works to make use of both the GAN-trained generator and discriminator to better detect anomalies based on both residual and discrimination losses.

Our contributions of this paper are summarized as follows: i), a novel GAN-based unsupervised anomaly detection method is proposed to detect anomalies (cyber-attacks) for complex multi-process cyber-phsyical systems with networked sensors and actuators; ii), the GAN model is trained with multiple time series, which adapts GAN from the image generation domain for time series generation by adopting the Long Short Term-Recurrent Neural Networks (LSTM-RNN) to capture the temporal dependency; iii), normal sequences with high dimension is uniformly utilized to train the GAN model to discriminate fake from real and reconstruct testing sequences from specific latent space simultaneously; iv), the discrimination loss calculated by the trained discriminator and the residual loss between reconstructed and real testing sequences (to make use of both the trained discriminator and generator) are combined together to detect anomalous points in the high dimensional time series, and the proposed method is shown to outperform existing methods in detecting anomalies due to cyber attacks in a complex  Secure Water Treatment (SWaT) system with six stages \cite{Jonathan2016}.

\section{Anomaly Detection with Generative Adversarial Training}
\label{sec:Model}
\subsection{GAN with LSTM-RNN}
\label{subsec:GAN}

Long Short Term-Recurrent Neural Networks (LSTM-RNN) had been shown  to be capable of learning complex time series by taking the information in backward (or even forward) time steps with memorise cells. In this work, in order to handle time series data of the CPS, both the generator ($G$) and discriminator ($D$) of GAN are substituted by LSTM-RNN. Following the architecture of a regular GAN framework \cite{Goodfellow2014}, the GAN model is trained as a two-player minimax game.
\begin{equation}
\begin{array}{lcl}
\label{eq:gan_minmax}
\min\limits_G \max\limits_D V(D,G) = \mathcal{E}_{x\backsim p_{data}(x)}\left[ \log D(x)\right]\\
\;\;\;\;\;\;\;\;\;\;\;\;\;\;\;\;\;\;\;\;\;\;+ \mathcal{E}_{z\backsim p_z(Z)}\left[\log (1-D(G(z)))\right]
\end{array}
\end{equation}

Specifically, the generator $G$, a LSTM-RNN model, implicitly defines a probability distribution for the generated samples, which can be written as $G_{rnn}(z)$, where $z$ is a distribution from the random latent space. The discriminator, which is another LSTM-RNN model, is then trained to minimise the average negative cross entropy between its predictions and sequence labels (e.g., train $D$ to recognize as many training samples as real as possible, and recognize as many generated samples as false as possible). Thus, the discriminator loss is
\begin{equation}
\label{eq:D_loss}
\begin{array}{lll}
D_{loss}=\frac{1}{m}\sum_{i=1}^{m}\left[\log D_{rnn}(x_i)+\log (1-D_{rnn}(G_{rnn}(z_i)))\right]\\
\Leftrightarrow \min \frac{1}{m}\sum_{i=1}^{m}\left[-\log D_{rnn}(x_i)-\log (1-D_{rnn}(G_{rnn}(z_i)))\right]
\end{array}
\end{equation}
where $x_i, i=1,...,m$ is the training samples which should be recognized as real, and $G_{rnn}(z_i), i=1,...,m$ is the generated samples that should be recognized as false. 

At the same time, the generator is trained to confuse the discriminator so that the discriminator would recognize as many generated samples as real as possible. In other words, the generator loss is
\begin{equation}
\label{eq:G_loss}
\begin{array}{lll}
G_{loss}=\sum_{i=1}^{m}\log (1-D_{rnn}(G_{rnn}(z_i)))\\
\Leftrightarrow \min \sum_{i=1}^{m}\log (-D_{rnn}(G_{rnn}(z_i)))
\end{array}
\end{equation}

The generator loss and discriminator loss are jointly dealt with by optimizers and used to update the parameters for $G_{rnn}$ and $D_{rnn}$.

\begin{algorithm}
\begin{algorithmic}
  \LOOP 
   \IF {epoch within number of training iterations}
    \FOR {the kth epoch}
     \STATE Generate samples from the latent space: 
     \STATE $Z=\lbrace z_i, i=1,...,m\rbrace \Rightarrow G_{rnn}(Z)$
     \STATE Conduct discrimination: 
     \STATE $X=\lbrace x_i, i=1,...,m\rbrace \Rightarrow D_{rnn}(X) $  
     \STATE $G_{rnn}(Z) \Rightarrow D_{rnn}(G_{rnn}(Z))$
     \STATE Update discriminator parameters by minimizing(descending) $D_{loss}$:
        \STATE $\min \frac{1}{m}\sum_{i=1}^{m}(-\log D_{rnn}(x_i)$
            \STATE $-\log (1-D_{rnn}(G_{rnn}(z_i))))$
     \STATE Update discriminator parameters by minimizing(descending) $G_loss$ :
        \STATE $\min \sum_{i=1}^{m}\log (-D_{rnn}(G_{rnn}(z_i)))$
     \STATE Record parameters of the discriminator and generator in the current iteration.    
    \ENDFOR   
   \ENDIF
   \FOR {the lth iteration}
    \STATE Mapping testing data back to latent space:
       \STATE $Z^k=\min\limits_{Z} Er(X^{tes}, G_{rnn}(Z^i))$
   \ENDFOR
   \STATE Calculate the residuals:
       \STATE $Res = \mid X^{tes}- G_{rnn}(Z^k)\mid$
   \STATE Calculate the discrimination results:
       \STATE $Pro=D_{rnn}(X^{tes})$
   \STATE Obtain anomaly score:
       \STATE $S=Res+Pro$
  \ENDLOOP
\end{algorithmic}
\caption{LSTM-RNN-GAN-based Anomaly Detection Strategy}
\label{algo:lstm-rnn-gan}
\end{algorithm}

\subsection{GAN-based Anomaly Score}
\label{subsec:AS}
As a newly arisen unsupervised learning method, both the generator and generator of GAN are jointly trained to represent the normal anatomical variability which is helpful for identifying anomalies. To make full use of the GAN model, both the trained generator and discriminator should be driven to make contributions to the anomaly detection. Following the formulation in \cite{Schlegl2017}, the anomaly detection for CPSs time series data consists of the following two parts.
\begin{itemize}
\item \textbf{Anomaly Detection with Discrimination}\\
Intuitively, the trained discriminator $D$ (after a sufficient number of iterations of adversarial training) is a direct tool for anomaly detection since it can distinguish fake from real with high sensitivity.
\item \textbf{Anomaly Detection with Residuals}\\
As mentioned in previous sections, the trained generator $G$, which is capable of generating realistic samples, is actually a mapping from the latent space to real data space: $G(Z):Z\rightarrow X$, and can be viewed as an inexplicit system model that reflects the normal data's distribution. Due to the smooth transitions of latent space mentioned in \cite{Radford2015}, the generator outputs similar samples if the inputs in the latent space are close. Thus, if it is possible to find the corresponding $Z^k$ in the latent space for the testing data $X^{tes}$, the similarity between $X^{tes}$ and $G(Z^k)$ could explain to which extent is $X^{tes}$ follows the distribution reflected by $G$. That is to say, residuals between $X^{tes}$ and $G(Z^k)$ could be utilized for identifying anomalies in testing data.\\
\end{itemize}

As shown in the right part of Fig. \ref{fig:GAN-CPS}, to find the optimal $Z^k$ that corresponds to the testing samples, we first sample a random set $Z^1$ from the latent space and obtain reconstructed raw samples $G(Z^1)$ by feeding it to the generator. Then, the samples from the latent space could be updated with the gradients obtained from the error function defined with $X^{tes}$ and $G(Z)$.
\begin{equation}
\label{eq:invLoss}
\min\limits_{Z^k} Er(X^{tes}, G_{rnn}(Z^k))=1-Simi(X^{tes},G_{rnn}(Z^k))
\end{equation} 
where the similarity between sequences could be defined as covariance for simplicity.

If after enough iteration rounds the error  is small enough, the samples $Z^k$ is recorded as the corresponding mapping in the latent space for the testing samples. Thus, the residual at time $t$ for testing samples is calculated as
\begin{equation}
\label{eq:residual}
Res(X^{tes}_t)=\sum_{i=1}^n \mid x_t^{tes,i}-G_{rnn}(Z^{k,i}_t) \mid 
\end{equation}
where $X^{tes}_t\subseteq \mathcal{R}^n$ is the measurements at time step $t$ for $n$ variables. In summary, the the anomaly score for anomaly detection is
\begin{equation}
\label{eq:score}
S_t^{tes}=\lambda Res(X^{tes}_t)+(1-\lambda)D_{rnn}(X^{tes}_t)
\end{equation}

Our GAN-based unsupervised anomaly detection strategy is summarized in Algo. \ref{algo:lstm-rnn-gan}. Mini-batch stochastic optimization based on Adam Optimizer and Gradient Descent Optimizer is used for updating the model parameters.

\subsection{Anomaly Detection Framework}
\label{subsec:AD}

We formulate the anomaly detection problem for multivariate time series as follows. First, consider an $m$-dimensional time series $X = \lbrace x^{(t)}, t=1,...,T\rbrace$ with length $T$ \footnote{Usually, $T$ should not be too large with purpose of monitoring and anmaly detection.}, where $x^{(t)}\in \mathcal{R}^m$ is an m-dimensional vector of readings for $m$ variables at time-instance $t$. Usually, in industry process or mechanical systems (such as the SWaT system considered in this paper), sensor measurements are large time-series with length $\mathcal{T}$ ($\mathcal{T}>>T$). Thus, multiple predefined time-series, $\mathbf{X}=\lbrace X^{(1),...,X^{(L)}}\rbrace$, can be obtained by taking a window of length $T$ over the raw data streams. The GAN model is trained based on the normal time-series dataset $\mathbf{X}^{real}$, and generates "fake" samples $\mathbf{X}^{gs}$ that "look real". Next, the testing time-series dataset $\mathbf{X}^{att}$ (or $\mathbf{X}^{tes}$), which is real-time CPSs data, can be analysed by the trained model to detect anomalous slots. 

However, the use of LSTM-RNN with high-dimensional inputs ( $\mathbf{X} \subseteq \mathcal{R}^{51}$ in the SWaT case) incurs higher computational cost than usual deep neural networks. Thus, in this paper, we adopt the Principal Component Analysis (PCA) to project the high-dimensional data into a PC projection space before feeding the data to the GAN model: $\mathbf{X}^{tes}\subseteq \mathcal{R}^m \Rightarrow \mathcal{X} ^{tes}\subseteq \mathcal{R}^n$. The projection is 
\begin{equation}
\label{eq:pca}
\begin{array}{lll}
P=PCA(\mathbf{X}^{real} )\\
\mathcal{X}^{tes}=\mathbf{X}^{tes}P^T
\end{array}
\end{equation}
where $\mathbf{X}^{tes}\subseteq \mathcal{R}^m$, $\mathcal{X} ^{tes}\subseteq \mathcal{R}^n$, $P\subseteq \mathcal{R}^{n\times n}$, $m$ is the original dimension (namely the number of system variables) and $n$ is number of reserved principal components. Then the projected variables are fed to the GAN-AD model and output anomaly scores according to Eq. (\ref{eq:score}). Next, the following label assigning function could be applied to identify whether the $i^{th}$ variable of  the testing time-series set $\mathcal{X}^{tes}$ at time step $t$ is being attacked or not.
\begin{equation}
\label{eq:dis_at}
\begin{array}{lll}
A_t^{tes,i}=\left\{\begin{matrix}
1, & if\; H\left(S(x_t^{tes,i}),1\right)>\tau\\
0, & else
\end{matrix}\right.
\end{array}
\end{equation}
where $t=1,...,T,\;i=1,...,n$. An anomaly is detected if the cross entropy error $H\left(.,.\right)$ for the anomaly score is higher than a certain value $\tau$. 

\section{SWaT System and Cyber-attacks}
\label{sec:CPSandAtt}

\subsection{Water Treatment System}
\label{subsec:CPS}

The Secure Water Treatment (SWaT) system is an operational test-bed for water treatment that represents a small-scale version of a large modern water treatment plant found in large cities \cite{Mathur2016} \footnote{The overall testbed design was coordinated with Singapore's Public Utility Board, the nation-wide water utility company, and constructed by a third party vendor. That collaboration ensured that the overall physical process and control system closely resemble real systems in the field, so that the results can be applied to real systems as well. For more information, please refer to https://itrust.sutd.edu.sg/research/testbeds/secure-water-treatment-swat/}. 

The water purification process in SWaT is composed of six sub-processes referred to as $P1$ through $P6$ \cite{Jonathan2016}. The first process is for raw water supply and storage, and $P2$ is for pre-treatment where the water quality is assessed. Undesired materials are them removed by ultra-filtration (UF) backwash in $P3$. The remaining chorine is destroyed in the Dechlorination process ($P4$). Subsequently, the water from $P_4$ is pumped into the Reverse Osmosis (RO) system ($P5$) to reduce inorganic impurities. Lastly, $P6$ stores the water ready for distribution. 

\subsection{Cyber-Attacks}
\label{subsec:Atta}

\begin{table}
  \caption{List of Cyber-attacks Inserted to the SWaT System}
  \label{tab:attacks}
  \centering
  \newcommand{\cc}[1]{\multicolumn{1}{c}{#1}}
  \begin{tabular}{llll}
  \hline\hline
          Process& Type& Attacked sensors& Attack Actuators\\
          \multirow{2}{*}{P1}&SSSP& LIT-101& MV-101; P-101; P-102\\
                             &SSMP& \multicolumn{2}{l}{(LIT-101 and MV-101)}\\  
          \hline  
          \multirow{3}{*}{P2}&SSSP& AIT-202& \\
                             &SSMP& \multicolumn{2}{l}{(P-203 and P-205)}\\  
                             &SSMP& \multicolumn{2}{l}{(P-201, P-203 and P-205)}\\  
          \hline
          \multirow{1}{*}{P3}&SSSP& LIT-301; DPIT-301& MV-303;MV-303;\\
                             &    &                  & MV-304; P-302\\
          \hline 
          \multirow{1}{*}{P4}&SSSP& LIT-401; FIT-401& \\
          \hline  
          \multirow{2}{*}{P5}&SSSP& AIT-504& MV-504\\
                             &SSMP& \multicolumn{2}{l}{(P-501 and FIT-502)}\\  
          \hline 
          \multirow{8}{*}{P1-6}&MSMP& \multicolumn{2}{l}{(UV-401, AIT-502 and P501)}\\  
                               &MSMP& \multicolumn{2}{l}{(P-602, DIT-301 MV-301)}\\
                               &MSMP& \multicolumn{2}{l}{(P-302 and LIT-401)}\\ 
                               &MSMP& \multicolumn{2}{l}{(LIT-101, P-101 and MV-201)}\\ 
                               &MSSP& \multicolumn{2}{l}{(AIT-402 and AIT-502)}\\
                               &MSSP& \multicolumn{2}{l}{(FIT-401 and AIT-502)}\\
                               &MSSP& \multicolumn{2}{l}{(P-101 and LIT-301)}\\
          \hline
          \multicolumn{4}{l}{*FIT-flower meter; LIT-water level transmitter}\\
          \multicolumn{4}{l}{*MV-motorized valve}\\
          \multicolumn{4}{l}{*P-water pump/dosing pump/Sodium bi-sulphate pump}\\
          \multicolumn{4}{l}{*AIT-chemical analyser; UV-dechlorinator meter }\\
          \multicolumn{4}{l}{*DPIT-differential pressure indicating transmitter}\\
          \multicolumn{4}{l}{*SSSP: single stage single point attack}\\
          \multicolumn{4}{l}{*SSMP: single stage multi point attack}\\
          \multicolumn{4}{l}{*MSMP: multi stage multi point attack}\\
          \multicolumn{4}{l}{*MSSP: multi stage single point attack}\\
  \hline\hline
  \end{tabular}
\end{table}

Various experiments have been conducted on the SWaT system to investigate cyber-attacks and respective system responses. Please refer to the SWat  website \footnote{http://itrust.sutd.edu.sg/research/dataset} for a detailed description of the attacks. A total of $36$ attacks were launched during the 2016 SWaT data collection process \cite{Jonathan2016}. Generally, the attacked points include sensors (e.g., water level sensors, flow-rate meter, etc.) and actuators (e.g., valve, pump, etc.). The summary of attacked points based on attack location and type is shown in Table \ref{tab:attacks}.  

As a test-bed for research in the area of cyber security, several related works have been published based on the SWaT dataset. Some of them focused on special attacks. For example, a distributed detection method for single stage multiple points attacks via system specific physical invariants is proposed in \cite{Adepu2016}. Also, Jonathan et al proposed to find attacks for the first process via RNN prediction and CUSUM detection \cite{Jonathan2017}. A model-based method, which derives a Kalman filter, was applied to estimate the evolution of the system dynamics on single variable basis \cite{Ahmed2017}.  In this work, we  consider all the aforementioned cyber-attacks as anomalous working conditions and train our proposed GAN-AD method to detect these anomalies for all the six processes in SWaT (results are shown in Section \ref{subsec:ADRes}). 

\subsection{SWaT Dataset}
\label{subsec:Dataset}
The 2016 SWaT data collection process lasted for a 11 days with the system  operated 24 hours per day. Various cyber-attacks were implemented on the testbed with different intents and divergent lasting durations (from a few minutes to an hour) in the final four days. The system was either allowed to reach its normal operating state before another attack was launched or the attacks were launched consecutively. The 2016 SWaT dataset and its associated attacks have the following characteristics \footnote{The raw data are not plotted in this paper due to page limit---please refer to \cite{Jonathan2016} for more information.}
 
\begin{itemize}
\item Different attacks may last for different time durations due to different scenarios.   Some attacks do not even take effect immediately.  The system stabilization durations also vary across attacks. Simpler attacks, such as those aiming at changing flow rates, require less time for the system to stabilize while the attacks that caused stronger effects on the dynamics of system  will require more time for stabilization. 
\item Attacks on one sensor (or actuator) may affect the performance on other sensors (or actuators), usually after a certain time delay.
\item Furthermore, similar types of sensors (or actuators) tend to respond to attacks in a similar fashion. For example,  attacks on the LIT101 sensor (a water level sensor in process 1) cause abnormal spikes in both LIT101 and LIT 301 (another water level sensor in process 3) but do not affect the readings of other sensors and actuators (such as flow rate sensor and power meter) in the system.
\end{itemize}

The aforementioned observations suggest we should take a multivariate approach in the modelling instead of taking each sensor or actuator in the CPS as an independent data source (univariate approach).  The underlying correlations between the sensors and actuators could be exploited to better detect anomalies in the system.

\section{Experiments}
\label{sec:ExpandRes}

\subsection{Data Pre-processing}
\label{subsec:Setup}

In the 2016 SWaT dataset, $51$ variables (sensor readings and actuator states) were measured for 11-days. Within the raw data, $496,800$ samples were collected under normal working conditions (data collected in the first 7-days), and $449,919$ samples were collected when various cyber-attacks were inserted to the system.  We eliminate the first $21,600$ samples from the training dataset since it took 5-6 hours to reach stabilization when the system was first turned on, according to \cite{Jonathan2016}.

We subdivide the original long multiple sequences into smaller time series by taking window across raw streams. Since the SWaT data were recorded every second, we set the window length  as $T$=$120$ (i.e. data collected within $2$ minutes). To capture the relevant dynamics of SWaT data, the window ($T$=$120$) is applied to the normal and testing datasets with shift length $SL_{nor}$=$10$ for normal dataset and $SL_{att}$=$120$ for testing dataset respectively.  In order to speed up the GAN training process and avoid over-fitting, the samples were down-sampled to one measurement every $10$ seconds by taking the median value. As a result, we obtained $47,508$ training samples and $3,720$ testing samples with sequence length $L$=$12$.

\subsection{System Architecture}
For this study,  we used an LSTM network with depth 3 and 100 hidden (internal) units for the generator. The LSTM network for the discriminator is relatively simpler with 100 hidden units and depth 1. Inspired by the discussion about latent space dimension in \cite{Esteban2017}, we also tried different dimensions and found that  higher latent space dimension generally generates better samples especially when generating multivariate sequences. Thus, we set the dimension of latent space as 15 in this study.

\subsection{Sample Generation}
\label{subsec:Evalu}

\begin{figure*}[t]
\centering
\begin{minipage}{0.3\textwidth}
  \centering
\includegraphics[width=1.0\textwidth]{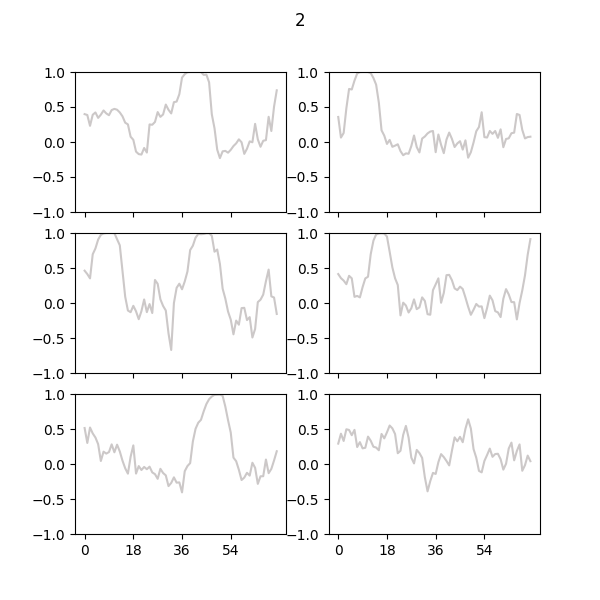}
\vspace{-0.9cm}
\subcaption[]{Generated samples at iteration=2.}\label{fig:GAN-VS-sub1}
\end{minipage}%
\begin{minipage}{0.3\textwidth}
  \centering
\includegraphics[width=1.0\textwidth]{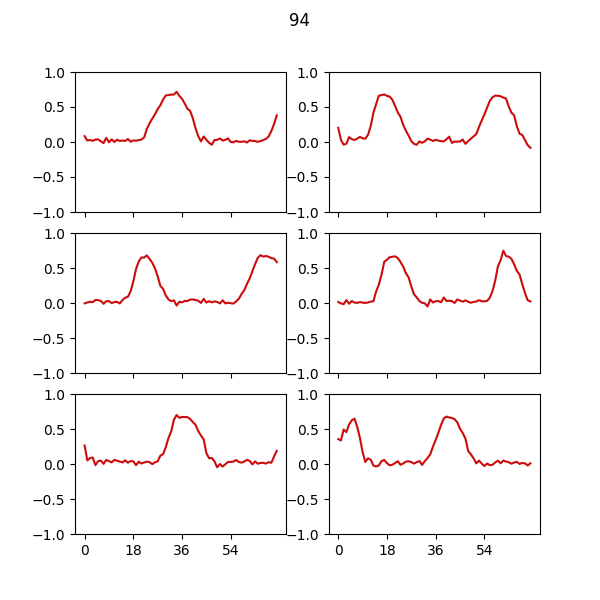}
\vspace{-0.9cm}
\subcaption[]{Generated samples at iteration=95.}\label{fig:GAN-VS-sub2}
\end{minipage}%
\begin{minipage}{0.3\textwidth}
  \centering
\includegraphics[width=1.0\textwidth]{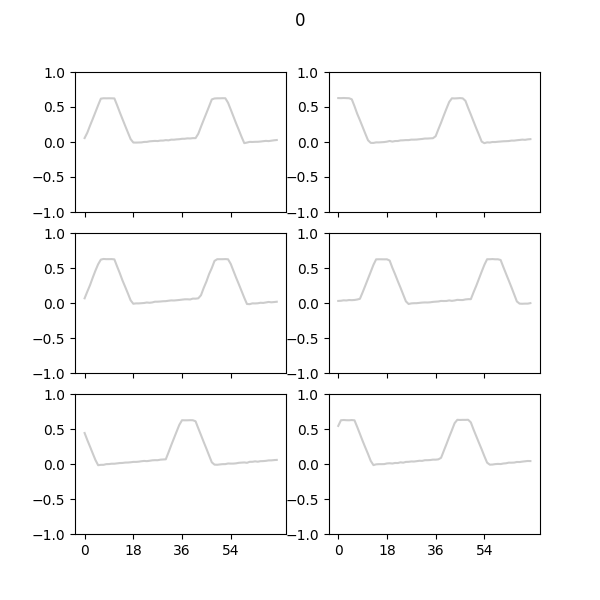}
\vspace{-0.9cm}
\subcaption[]{Original samples.}\label{fig:GAN-VS-sub3}
\end{minipage}%
\vspace{-0.1cm}
\caption{Comparison between generated samples at different traning stages:  GAN-generated samples at early stage are quite random while those generated at later stages almost perfectly took the distribution of original samples.} \label{fig:GAN-vs}
\end{figure*}

First, we visualize the data samples generated by our GAN versus the actual samples from the CPS.  As can be observed in Fig. \ref{fig:GAN-vs},  our GAN generated samples that were clearly different from the training data in the early learning stage.  However, after sufficient number of iterations,  the generator is able to output realistic samples for the various sensors and actuators in the system. 

\begin{figure}[h]
\centering
\includegraphics[width=0.5\textwidth, height=3cm]{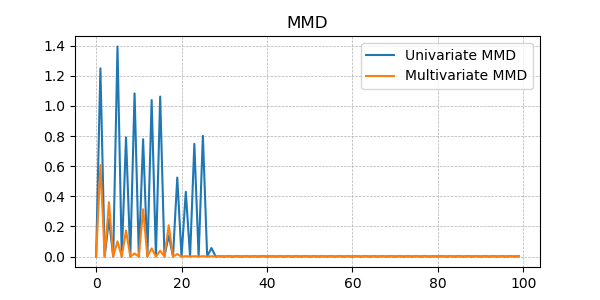}
\vspace{-3mm}
\caption{MMD: generation for multiple time series v.s. generation single time series.} 
\label{fig:mmd}
\end{figure}

\begin{figure*}[t]
\centering
\begin{minipage}{0.45\textwidth}
  \centering
\includegraphics[width=1.0\textwidth]{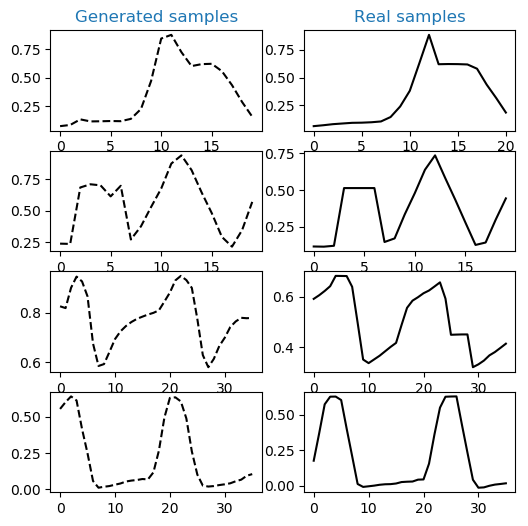}
\vspace{-0.5cm}
\subcaption[]{Univariate.}\label{fig:GS_RSI}
\end{minipage}%
\begin{minipage}{0.45\textwidth}
  \centering
\includegraphics[width=1.0\textwidth]{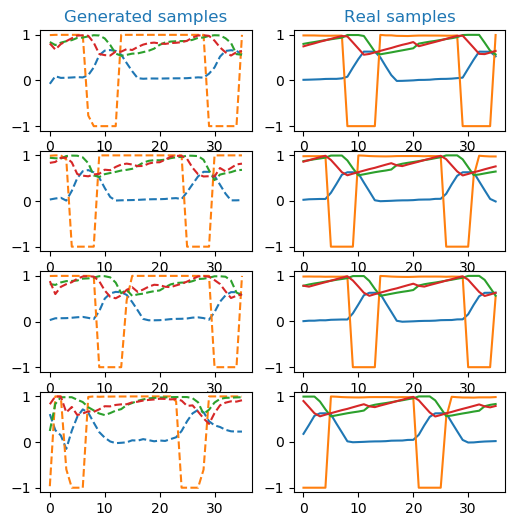}
\vspace{-0.5cm}
\subcaption[]{Multivariate.}\label{fig:GS_RSII}
\end{minipage}%
\vspace{-0.1cm}
\caption{Visualization of generated samples and original samples for both univariate and multivariate cases. Note that in the multivariate case 4 variables are fed to the GAN model simultaneously.} 
\label{fig:GSvsOS}
\end{figure*}

We use Maximum Mean Discrepancy (MMD) to evaluate whether the GAN model has learned the distributions of the training data.  MMD  is one of the training objectives for moment matching networks. 
\begin{equation}
\label{eq:mmd}
\begin{array}{lll}
MMD(Z_j,X_{tes})=\frac{1}{n(n-1)}\sum_{i=1}^n \sum_{j\neq i}^n K(Z_i^k, Z_j^k)\\
\;\;\;\;\;\;\;\;\;\;\;\;\;\;\;\;\;\;\;\;\;\;-\frac{2}{mn}\sum_{i=1}^n \sum_{j=1}^m K(Z_i^k, X_j^{tes})\\
\;\;\;\;\;\;\;\;\;\;\;\;\;\;\;\;\;\;\;\;\;\;+\frac{1}{m(m-1)}\sum_{i=1}^m \sum_{j\neq i}^m K(X_i^{tes}, X_j^{tes})\\
\end{array}
\end{equation}

We plot the MMD values across GAN training iterations for generating univariate samples and multivariate samples in Fig. \ref{fig:mmd}. In both cases, we can observe that the MMD values tend to converge to small values after 20-30 iterations. Interestingly,  the early MMD values of multivariate samples were lower than that of univariate samples, and the MMD for multivariate samples also converged faster than the univariate case. This suggests that using multiple data streams can help with the training of GAN model. 
Fig. \ref{fig:GSvsOS} shows the performance of our GAN in generating both univariate and multivariate samples.  As can be seen, after more than 50 iterations, the GAN model could  generate realistic time sequences even for the  multivariate case.

\subsection{Evaluation Metric}

We use the following metrics, namely Accuracy (Accu), Precision (Pre), Recall (Rec), $F_1$ score, and False Positive Rate (FPR) to evaluate the anomaly detection performance of GAN-AD.
\begin{equation}
\label{eq:Accu}
Accu=\frac{TP+TN}{TP+TN+FP+FN}
\end{equation}
\begin{equation}
\label{eq:Pre}
Pre=\frac{TP}{TP+FP}
\end{equation}
\begin{equation}
\label{eq:Rec}
Rec=\frac{TP}{TP+FN}
\end{equation}
\begin{equation}
\label{eq:F1}
F_{1}=2\times \frac{Pre \times Rec}{Pre+Rec}
\end{equation}
\begin{equation}
\label{eq:FPR}
FPR=\frac{FP}{FP+TN}
\end{equation}
where $TP$ is the correctly detected anomaly ($A_t=1$ while real label $L_t=1$), $FP$ is the falsely detected anomaly ($A_t=1$ while real label $L_t=0$), $TN$ is the correctly assigned normal ($A_t=0$ while real label $L_t=0$), and $FN$ is the falsely assigned normal ($A_t=0$ while real label $L_t=1$).

\begin{table}[!t]
  \caption{Anomaly Detection Rates for All Attacks by Checking Different Combinations of Measuring Points}
  \label{tab:detectionRes}
  \centering
  \newcommand{\cc}[1]{\multicolumn{1}{c}{#1}}
  \begin{tabular}{lllllll}
  \hline\hline
     Point& Method& Accu& Pre& Rec& $F_1$& FPR\\
    \hline
    \multirow{2}{*}{LIT-101}&CUSUM&     86.63& 36.42& 26.86& 0.30& 7.39\\
                            &GAN-AD&    87.63& 50.00& 1.75& 0.03& 9.32\\
    \hline
    \multirow{2}{*}{P-101}&CUSUM&     75.52& 24.62& 43.97& 0.31& 19.83\\
                          &GAN-AD&    80.72& 22.24& 15.43& 0.18& 8.71\\    
    \hline
    \multirow{2}{*}{AIT-202}&CUSUM&   55.67& 9.24& 25.15& 0.13& 39.45\\
                            &GAN-AD&  60.22& 4.58& 17.10& 0.07& 35.48\\    
    \hline
    \multirow{2}{*}{LIT-301}&CUSUM&    81.50& 12.92& 9.02& 0.10& 8.43\\
                            &GAN-AD&   86.85& 22.22& 1.04& 0.02& 0.52\\
    \hline
    \multirow{2}{*}{DPIT-301}&CUSUM&    84.13& 18.46& 17.14& 0.17& 8.41\\ 
                             &GAN-AD&   84.40& 25.00& 2.67& 0.05& 1.18\\
     \hline
    \multirow{2}{*}{MV-303}&CUSUM&    71.55& 9.67& 19.18& 0.12& 22.01\\ 
                             &GAN-AD& 87.68& 17.54& 3.00& 0.05& 1.74\\    
    \hline
    \multirow{2}{*}{LIT-401}&CUSUM&  88.28& 47.80& 58.53& 0.52& 7.99\\
                            &GAN-AD& 80.35& 11.68& 9.94& 0.10& 10.14\\
    \hline
    \multirow{2}{*}{FIT-401}&CUSUM&  12.90& 12.90& 100.00& 0.23& 100\\
                            &GAN-AD& 85.40& 39.36& 4.32& 0.08& 1.09\\
    \hline
    \multirow{2}{*}{AIT-504}&CUSUM& 70.97& 6.23& 14.38& 0.08& 23.01\\
                            &GAN-AD& 86.03& 14.74& 14.35& 0.14& 11.14\\    
    \hline
    \multirow{4}{*}{All}    &SPE$^1$&           87.24& 20.49& 2.25& 0.04& 1.19\\
                            &SPE$^5$&           82.81& 24.92& 21.63& 0.23& 8.87\\
                            &GAN-AD$^1$&        90.57& 85.71& 7.20& 0.13& \textbf{0.13}\\
                            &GAN-AD$^5$& \textbf{94.80} & \textbf{93.33}& \textbf{63.64}& \textbf{0.75}& 0.46\\
 
    \hline
    \multicolumn{7}{l}{$*^1$ means only one principal component is chosen }\\
    \multicolumn{7}{l}{$*^5$ means the first five principal components are chosen}\\
  \hline\hline
\end{tabular}
\end{table}

\subsection{Anomaly Detection Results}
\label{subsec:ADRes}
We evaluate the anomaly detection performance for both the univariate and  multivariate cases.  In the univariate case, we compare the performance against the CUSUM approach which was used in previous works such as \cite{Jonathan2017}.  For the multivariate case, we compare GAN-AD against PCA-based unsupervised detection by inspecting the Squared Predicted Error (SPE, i.e., residual distance).  The anomaly detection results and comparisons are summarized in Table \ref{tab:detectionRes}. We showed the results on $9$ variables (i.e. sensors/actuators) for discussion. The $9$ variables include sensors and actuators from different processes of the system.  They were also attack points for single-point attacks.   

To evaluate the major contribution of this work (i.e. detection of anomalies for multiple sequences), all the measured variables are fed uniformly into the GAN-AD model (as mentioned previously, to decrease the computational load, the 51-dimensional data is projected to a lower dimensional space with the help of PCA). Since PCA is also a popular unsupervised multivarite anomaly detection method by inspecting the Squared Predicted Error (SPE, i.e., residual distance) values, we also evaluate SPE-based detection against our proposed GAN-AD method.

\subsubsection{Univariate}
\label{subsubsec:Uni}

As shown in Table \ref{tab:detectionRes}, univariate detection generally perform poorly with high false positive rate, and low precision and recall. We also observed that:

\begin{itemize}
	\item The water level sensors (e.g., LIT-101, LIT-301 and LIT-401) have the best anomaly detection performance---both CUSUM and GAN-AD were able to achieve good detection accuracy rate (near or above $80\%$). This is because other than the direct attacks on those sensors, most of attacks applied to the other sensors (or actuators) also affected the water levels indirectly. 
	
	\item The accuracy values of AIT-504 were $70.97\%$ and $86.03\%$ (CUSUM and GAN-AD respectively) were better than its earlier counterpart AIT-202 ($55.67\%$ and $60.22\%$).  This was because the chemical accumulation in $P5$ (where inorganic impurities are reduced) is relevant for anomaly detection after the chorine has been added to the system in $P3$ and $P4$. $P2$ was just a pre-treatment process where the quality of raw water for $P1$ is assessed.
	
	\item Detection results with actuators P-101 and MV-303 were not as poor as expected. Despite of the $0/1$ values for on/off (open/close) states, both CUSUM and GAN-AD managed to achieve accuracy of up to $71.55\%-75.52\%$ and $80.72\%-87.68\%$ respectively. While the $0/1$ states do not provide variance across numerical dimension, the frequency of how often "0" and "1" appears along the time line was still useful for anomaly detection.
	
	\item For the water flow-rate sensor FIT-401, the results by CUSUM was extremely poor while  GAN-AD  performed well. One possible reason for CUSUM's $100\%$ recall and false positive rate was that CUSUM did not assign any negative labels to the testing samples (i.e., recognizing all samples as anomalies) due to unsuitable normal ranges. On looking closely at the normalized raw data, we observed that the values of flow rate meter  took a roughly $0/1$ shape (just like the actuator states). However, the flow rate at the high points were not static $1$s but varied with high frequency. CUSUM is unable to capture such "bi-variant" characteristic  and hence performed badly in this case.  On the other hand,  GAN-AD was able to handle this and generated acceptable accuracy rate.
	
\end{itemize}

\subsubsection{Multivariate}

A key contribution of this work is applying our proposed GAN-AD method to solve the multivariate anomaly detection problem for time series data. For dimensional reduction, instead of directly feeding the high dimensional data to the GAN-AD model, we used PCA  to project the raw data into a lower dimensional principal space, as described in Eq. (\ref{eq:pca}). 

We plot the variance rate of the first 10 Principal Components (PC) in Fig. \ref{fig:pca}. As shown in the figure, there is one main PC that explained more than $50\%$ of the variance for the SWaT data.  Also, the PCs after the $5^{th}$ one contribute little to the overall variance (near to $0$). As such, we projected the SWaT data to the most variant PC (the first one) as well as the first $5$, and then applied the GAN-AD to detect anomalies for the projected data. For comparison, we also performed standard PCA-based anomaly detection  by inspecting the testing dataset with the Squared Predicted Error (SPE, i.e., the residual distances calculated by PCA projection) method. 

The bottom part of Table \ref{tab:detectionRes} shows the performance of  multivariate anomaly detection using SPE and our proposed GAN-AD. The results showed  about $3\%-12\%$ improvement with the proposed GAN-AD. The GAN-AD method also achieved $50\%$-$60\%$ higher precision and $5\%$-$40\%$ higher recall compared with SPE by assigning more true positives (correctly detected anomalies).

We also compared both GAN-AD and SPE based on PC=1 and PC=5. That is, we conducted SPE with the first one and five principal components, while the raw data were projected to the first one and five principal components correspondingly before being fed into the GAN-AD. It is interesting to see that for both GAN-AD and SPE the recall rates with PC=5 (hold more than $90\%$ variate rate) were obviously higher that with PC=1 (which only contains around $50\%$ variate rate as shown in  Fig. \ref{fig:pca}), which implies that using more principal components could reduce false negatives.

In terms of false positive rates, GAN-AD with PC=1 achieved the best FPR amongst all (both univariate and multivariate). Although GAN-AD with PC=5 outperformed others in the aspect of accuracy, precision, recall, and $F_1$, its false positive rate was slightly higher than that by GAN-AD with PC=1. Similar phenomenon could be observed in multivariate detection by SPE. This indicates that the improvement of detection accuracy (as well as precision and recall) was built upon the sacrifice of more false positives due to the noisy information brought in by adding four more less important PC dimensions. 

The results in Table \ref{tab:detectionRes} also showed that:

\begin{figure}[!h]
\centering
\includegraphics[width=0.5\textwidth, height=3cm]{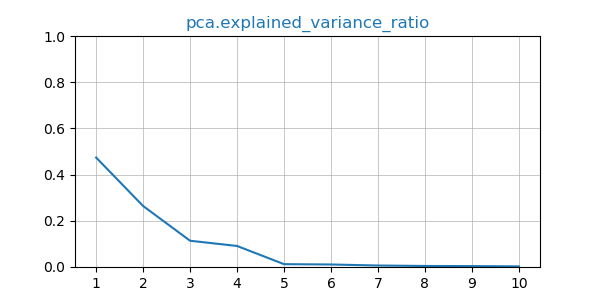}
\vspace{-5mm}
\caption{Variance Ratio of Principal Component for the SWaT data.} 
\label{fig:pca}
\end{figure}

\begin{itemize}
	\item Generally, the univariate detection cannot compete with multivariate detection. To be specific, univariate detection results in widespread low precision and recall (note that multivariate detection by SPE also performed poorly in terms of these two factors, which we will discuss in the next point), and high FPR. This observation demonstrates the multivariate detection is applicable for complex CPSs with interconnected IoT of sensors and actuators generating large amounts of time series. 
	
	\item In fact,  even the baseline multivariate anomaly detection method SPE can compete with most of the univariate detection results in terms of accuracy,. However, SPE does not compare well with univariate detection for LIT-101, DPIT-301, LIT-401 and FIT-301 in terms of precision. As noisy information can be accumulated when simply projecting the whole raw dimensions,  it can sometimes cause negative effect in assigning true positives \cite{Dan4:2017}. One possible future work would be to consider selecting suitable variables and appointing different weights according to their importance levels, instead of treating all the variables uniformly in one plain framework as in the current study.
	
\end{itemize}

\section{Conclusions}
\label{sec:ConcandFW}

Cyber-Physical Systems are large, complex, and affixed with networked sensors and actuators that generate large amounts of data streams. These data streams and their underlying system dependencies can potentially be mined for dynamic detection of possible intrusion incidents. In this paper, we have explored the use of GAN to simultaneously train a deep learning network to model the distributions of multi-sensor data streams in a CPS under normal operating conditions, and another to detect anomalies due to cyber attacks being carried out against the CPS in an unsupervised fashion. We have proposed a novel GAN-based Anomaly Detection (GAN-AD) method that directly utilizes both the discriminator and the generator trained on multivariate time series to detect anomalies. We have tested our approach on a complex CPS dataset from a Secure Water Treatment Testbed (SWaT) and showed that the proposed GAN-AD was able to outperform existing unsupervised detection methods.
 
For future work, we will explore the use of GAN-AD for other IoT applications such as predictive maintenance and fault diagnosis for smart buildings and machineries. In terms of the GAN-AD methodology, instead of simply feeding multiple sequences uniformly into a fully connected network, we plan to enhance GAN-AD with a multi-GAN framework to better capture the extrinsic knowledge about relationships among the networked sensors and components. We will also conduct further research on feature selection for multivariate anomaly detection, and investigate principled methods for choosing the latent dimension and PC dimension with theoretical guarantees.


\section*{Acknowledgement}
This material is based on research/work supported by the Singapore National Research Foundation and the Cybersecurity R\&D Consortium Grant Office under Seed Grant Award No.CRDCG2017-S05.

%

\ifCLASSOPTIONcaptionsoff
  \newpage
\fi



%
%
%

\bibliographystyle{IEEEtran}
\bibliography{GANReference}

%

%
%
%




\end{document}